\PassOptionsToPackage{unicode}{hyperref}
\PassOptionsToPackage{hyphens}{url}
\PassOptionsToPackage{dvipsnames,svgnames,x11names}{xcolor}
\documentclass[
  11pt,
]{article}
\usepackage{xcolor}
\usepackage[margin=1in]{geometry}
\usepackage{amsmath,amssymb}
\usepackage{iftex}
\ifPDFTeX
  \usepackage[T1]{fontenc}
  \usepackage[utf8]{inputenc}
  \usepackage{textcomp} 
\else 
  \usepackage{unicode-math} 
  \defaultfontfeatures{Scale=MatchLowercase}
  \defaultfontfeatures[\rmfamily]{Ligatures=TeX,Scale=1}
\fi
\usepackage{lmodern}
\ifPDFTeX\else
\fi
\IfFileExists{upquote.sty}{\usepackage{upquote}}{}
\IfFileExists{microtype.sty}{
  \usepackage[]{microtype}
  \UseMicrotypeSet[protrusion]{basicmath} 
}{}
\makeatletter
\@ifundefined{KOMAClassName}{
  \IfFileExists{parskip.sty}{%
    \usepackage{parskip}
  }{
    \setlength{\parindent}{0pt}
    \setlength{\parskip}{6pt plus 2pt minus 1pt}}
}{
  \KOMAoptions{parskip=half}}
\makeatother
\usepackage{longtable,booktabs,array}
\usepackage{caption}
\usepackage{calc} 
\usepackage{etoolbox}
\makeatletter
\patchcmd\longtable{\par}{\if@noskipsec\mbox{}\fi\par}{}{}
\makeatother
\IfFileExists{footnotehyper.sty}{\usepackage{footnotehyper}}{\usepackage{footnote}}
\makesavenoteenv{longtable}
\usepackage{graphicx}
\makeatletter
\newsavebox\pandoc@box
\newcommand*\pandocbounded[1]{
  \sbox\pandoc@box{#1}%
  \Gscale@div\@tempa{\textheight}{\dimexpr\ht\pandoc@box+\dp\pandoc@box\relax}%
  \Gscale@div\@tempb{\linewidth}{\wd\pandoc@box}%
  \ifdim\@tempb\p@<\@tempa\p@\let\@tempa\@tempb\fi
  \ifdim\@tempa\p@<\p@\scalebox{\@tempa}{\usebox\pandoc@box}%
  \else\usebox{\pandoc@box}%
  \fi%
}
\def\fps@figure{htbp}
\makeatother
\usepackage[]{natbib}
\usepackage{needspace}
\usepackage{bookmark}
\IfFileExists{xurl.sty}{\usepackage{xurl}}{} 
\makeatletter
\@ifundefined{xmpquote}{}{}
\makeatother
\hypersetup{
  pdftitle={Unknown is not normal: separating language-model extraction from rule-based decision logic for clinical risk scores},
  pdfauthor={Nicolás Vera Zúñiga --- Independent researcher, Chile --- nicovera@quetru.cl},
  colorlinks=true,
  linkcolor={blue},
  filecolor={Maroon},
  citecolor={blue},
  urlcolor={blue},
  pdfcreator={LaTeX via pandoc}}

\title{Unknown is not normal: separating language-model extraction from
rule-based decision logic for clinical risk scores}
\author{Nicol\'as Vera Z\'u\~niga\\
Independent Researcher, Chile\\
\texttt{nicovera@quetru.cl}}
\date{}
\date{}

\ifPDFTeX
\DeclareUnicodeCharacter{2265}{\ensuremath{\geq}}
\DeclareUnicodeCharacter{2264}{\ensuremath{\leq}}
\DeclareUnicodeCharacter{00D7}{\ensuremath{\times}}
\DeclareUnicodeCharacter{2192}{\ensuremath{\rightarrow}}
\DeclareUnicodeCharacter{00B5}{\ensuremath{\mu}}
\DeclareUnicodeCharacter{00B0}{\ensuremath{^{\circ}}}
\DeclareUnicodeCharacter{00B1}{\ensuremath{\pm}}
\DeclareUnicodeCharacter{2212}{\ensuremath{-}}
\fi
\begin{document}
\maketitle

\begin{abstract}
Large language models (LLMs) are increasingly used to compute clinical
risk scores from free-text notes. Notes are often incomplete, and
treating undocumented findings as normal can silently misclassify
patients. We test whether separating three-state extraction (present,
absent or unknown, by an LLM) from decision logic (deterministic code
computing score bounds over unknown inputs) lets a system ask only
questions that can change the decision. On 1,200 synthetic emergency
cases across six calculators (HEART, CURB-65, qSOFA, PERC, Wells,
Cockcroft-Gault), with a simulated clinician answering questions, we
compared this bounds policy with asking for every missing input, a
missing-equals-normal schema, and an end-to-end LLM agent (Claude Opus
5.5). With Claude Haiku 4.5 as extractor, the bounds policy matched
ask-all accuracy (99.4\% vs 99.4\%) with half the questions (0.92 vs
1.78 per case) and no irrelevant ones. Treating missing as normal
dropped accuracy to 91.2\% and under-triaged 8.5\% of patients (95\% CI
7.1--10.2), and under-triage persisted under messy notes and a noisy
clinician. The agent was equally accurate under ideal conditions
(99.6\%) but 9.5\% of its questions were irrelevant; with a noisy
clinician it was less accurate than the bounds policy (83.5\% vs 87.0\%,
p\textless0.001) and committed prematurely in 2.7\% of cases (bounds:
0\%). A 9B local model as extractor reached oracle-level accuracy
(99.8\%). In 584 real case reports from MedCalc-Bench, only 52\%
contained enough information to determine the category (HEART 13\%).
Routing decisions through code that reasons explicitly about unknowns
avoids premature commitment and irrelevant questions, halves the
questions asked, and works with small local models.
\end{abstract}

\section{1. Introduction}\label{introduction}

Clinical scores such as HEART (chest pain), CURB-65 (pneumonia), qSOFA
(suspected sepsis), PERC and Wells (pulmonary embolism) and
Cockcroft-Gault (renal dosing) are defined over a few inputs with
explicit thresholds. What matters clinically is usually the
\emph{decision category}, not the exact score: a HEART score of 0--3 is
low risk, 4--6 moderate, 7 or more high. LLMs are now used to compute
these scores from notes
\citep{khandekar2024medcalc, wang2025scores, zhu2026medmcpcalc}, and the
standard benchmark, MedCalc-Bench, treats an input the note does not
mention as absent or normal \citep{khandekar2024medcalc}.

That convention is not neutral. ``No mention of hemoptysis'' is not ``no
hemoptysis'', and a score computed as if it were can place a patient in
a lower-risk category. The safe behaviour is to recognise when the
documented facts do not determine the decision and then ask only for
facts that could change it. LLMs handle this poorly in both directions,
committing when the category is undetermined and abstaining when it is
determined \citep{watanabe2026clindet}, and interactive benchmarks show
that models gathering information ask too little or too much
\citep{li2024mediq, schmidgall2024agentclinic, johri2025craftmd}.

We test a simple division of labour (Figure 1). A language model reads
the note into three-valued facts (present with a value, explicitly
absent, or unknown), each with an exact evidence quote and a confidence.
Deterministic code computes the range of scores still possible given the
unknowns (the score's \emph{bounds}), decides whether the category is
determined, and if not asks only about unknowns that could change it. We
compare this with asking for every missing input, with the same pipeline
treating missing as normal, and with an end-to-end LLM agent. We
pre-specified four hypotheses: (H1) the bounds policy asks far fewer
questions than ask-all at equal accuracy; (H2) an end-to-end agent both
asks questions that cannot change the category and answers before the
category is determined; (H3) three-valued extraction reduces silent
missing-as-absent errors compared with a binary schema; (H4) confirming
low-confidence, decision-critical values catches extraction errors at a
small question cost.

\begin{figure}
\centering
\pandocbounded{\includegraphics[keepaspectratio,alt={Pipeline. A language model reads the note into three-valued facts with evidence and confidence. Code normalises units, computes the score's bounds over the unknowns and, if the decision category is not determined, asks the clinician only about inputs that could change it; each answer updates the bounds.}]{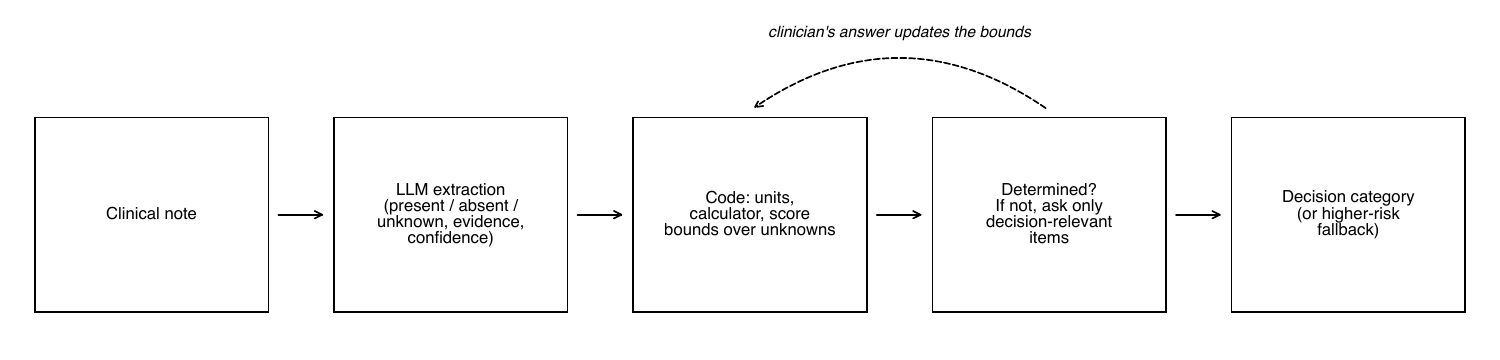}}
\caption{Pipeline. A language model reads the note into three-valued
facts with evidence and confidence. Code normalises units, computes the
score's bounds over the unknowns and, if the decision category is not
determined, asks the clinician only about inputs that could change it;
each answer updates the bounds.}
\end{figure}

\section{2. Methods}\label{methods}

\textbf{Calculators and bounds.} The six calculators were implemented
from their primary sources
\citep{six2008heart, lim2003curb65, seymour2016qsofa, singer2016sepsis3, kline2004perc, kline2008perc, wells2000pe, vanbelle2006christopher, cockcroft1976creatinine},
with decision categories HEART 0--3 / 4--6 / ≥7, CURB-65 0--1 / 2 / ≥3,
qSOFA ≥2 positive, PERC negative if all criteria are met, Wells PE
likely if \textgreater4, and Cockcroft-Gault \textless30 /
30--\textless60 / ≥60 mL/min. Where a source was ambiguous (for example,
HEART troponin bands or the BUN equivalent of CURB-65 urea) we
implemented a documented reading; the author, an emergency physician,
reviewed all readings (Supplementary S1). Units are converted by code,
never by the model. Inputs are boolean, ordinal or numeric; numeric
inputs are split into regions at the calculator's cut-points, and
achievable scores are enumerated exactly over the regions
(Cockcroft-Gault, which is monotone, is evaluated at interval corners).
The category is \emph{determined} when only one category is possible,
and an unknown input is \emph{decision-relevant} if its value changes
the category for some completion of the other unknowns. For example,
with 2 documented HEART points and troponin (0--2 points) unknown, the
possible scores 2--4 span low and moderate risk, so troponin is asked;
with 0 documented points they do not, and nothing is asked.
Property-based tests \citep{maciver2019hypothesis} checked that every
completion falls within the bounds and that inputs labelled irrelevant
never change the category.

\textbf{Synthetic cohort and notes.} For each calculator we sampled 200
patients from plausible emergency-department priors (Supplementary S2).
Each input was left undocumented with probability 0.3, and cases were
resampled so that half had a category the note did not determine;
results can be reweighted to the natural share (Supplementary S3). Cases
carried traps: negations, values in other units, values from a previous
visit, superseded readings, and comorbidities implied only by a
medication. Claude Sonnet 5 wrote each note from a fact sheet, stating
documented values exactly and omitting undocumented inputs, including
indirect cues. Every note passed rule checks and an independent LLM
judge (Claude Opus 5.5) comparing each input with the ground truth (22
of 1,200 needed one retry). A ``messy'' end-of-shift version (fragments,
abbreviations, typos, copied-forward text) was rendered for 600 cases;
559 passed validation and form the paired set for harder conditions.

\textbf{Extraction.} For every input the extractor returned a status, a
value, a unit and the shortest exact supporting quote, plus a
confidence. Code rejected any claim whose quote was not an exact
substring of the note (the input then became unknown), converted units,
and rejected implausible values. Extractors were an oracle returning the
documented facts (a perfect-extraction reference), Claude Haiku 4.5
(through headless Claude Code with tools disabled and schema-constrained
output; self-reported confidence), and Qwen3.5-9B \citep{qwen35} run
locally through Ollama \citep{ollama} (confidence from token
log-probabilities). Confidence was calibrated by temperature scaling
\citep{guo2017calibration} and isotonic regression
\citep{zadrozny2002isotonic} on a 30\% development split.

\textbf{Policies.} \emph{Ask-all} asks about every input left unknown.
\emph{Bounds} asks only decision-relevant unknowns, one at a time,
recomputing the bounds after each answer. \emph{Bounds + checks} adds
value-of-information ordering and asks the clinician to confirm
decision-critical values with calibrated confidence below 0.9.
\emph{Missing = normal} is the bounds policy with missing inputs treated
as normal, the MedCalc-Bench convention. \emph{Agent} is an end-to-end
agent: Claude Opus 5.5 receives the note and the input list and, turn by
turn, asks the clinician, runs the calculator, or answers (or declares
the case undeterminable). When no question can settle the category, a
policy abstains; for secondary analyses an abstention becomes the
highest-risk category still possible (for the agent, the calculator's
highest-risk category).

\textbf{Simulated clinician.} A deterministic simulator answered from
the hidden truth; troponin was not yet available in 10\% of cases. The
noisy clinician could not answer 10\% of questions, answered wrongly 5\%
of the time (a flipped yes/no, a graded input one level off, or a number
off by 10--25\%), and gave 25\% of numeric answers as a ±10\% range. The
noise is fixed per case and input, so all policies face the same
clinician.

\textbf{Outcomes and statistics.} Primary outcomes were
decision-category accuracy, with abstention counted as incorrect, and
questions per case. Secondary outcomes were under-triage (a lower-risk
category than the truth) and over-triage; premature commitment
(answering while the category was undetermined); irrelevant questions
(whose answer could not change the category); silent missing-as-absent
claims; extraction accuracy; and calibration. A comorbidity implied only
by a medication is not determined by the note, so ``present'' and
``unknown'' were both accepted there. Proportions have Wilson intervals
\citep{wilson1927probable} and mean questions bootstrap intervals.
Paired comparisons used exact McNemar \citep{mcnemar1947note} and
Wilcoxon signed-rank tests, Holm-adjusted \citep{holm1979simple} across
four pre-specified comparisons per condition.

\textbf{Real notes.} We used MedCalc-Bench Verified
\citep{khandekar2024medcalc} and its label audits
\citep{ye2025stewardship, krohngrimberghe2026audit} for the five
overlapping calculators: 100 test notes and a sample of 584 training
notes (published, de-identified case reports; CC-BY-SA 4.0). We report
whether our code reproduces the benchmark labels from its annotated
inputs, how often the note determines the category, whether the truth
stays within our bounds, and accuracy under the missing-equals-normal
convention.

\textbf{Implementation.} All runs cost \$0 in API fees (Claude through a
subscription; Qwen on a 16 GB laptop). Each run is one configuration
file with a fixed seed, and every model call is cached, so all tables
and figures regenerate without new model calls. Code:
https://github.com/nicoveraz/calc-bounds (MIT licence; archived at
https://doi.org/10.5281/zenodo.23004726).

\section{3. Results}\label{results}

\textbf{Ideal conditions (Table 1).} H1 was supported. The bounds policy
matched ask-all accuracy with every extractor while asking 44--48\%
fewer questions (Haiku: 99.4\% for both, 0.92 vs 1.78 questions;
p\textless0.001). About half of ask-all's questions could not change the
decision; the bounds policy asked none. The saving held at 10\%, 30\%
and 50\% missingness (34--49\% fewer questions; Supplementary S3). H3
was supported: treating missing as normal reduced accuracy to
91.2--91.8\% (p\textless0.001), answered before the category was
determined in 39--49\% of cases, and under-triaged 8.2--8.8\% of
patients, against 0.0--0.1\% for the bounds policy. Haiku itself
occasionally turned an undocumented input into ``absent'' (0.022 per
case), which explains most of its small accuracy loss. Qwen missed 17\%
of documented negatives but labelled them unknown, so they were asked
rather than assumed. H2 was partly supported: the agent was as accurate
(99.6\%) but asked more questions (0.99 per case, p\textless0.001),
9.5\% of them irrelevant, and answered prematurely in 0.2\% of cases. H4
was partly supported: confirmation raised Haiku accuracy from 99.4\% to
99.7\% with fewer questions overall (0.90 vs 0.92), but the gain was not
significant because Haiku made few errors to catch. Calibration reduced
expected calibration error to 0.002 for both extractors (Supplementary
S5).

\textbf{Harder conditions (Table 1, Figures 2--3).} A noisy clinician
cost every policy that asks 12--16 accuracy points, mostly through
abstention: a ``don't know'' to the one decisive question leaves the
category open. When they did answer, ask-all and the two bounds policies
were 96.8--97.8\% correct. The agent became less accurate than the
bounds policy (83.5\% vs 87.0\%, p\textless0.001) and committed
prematurely in 2.7\% of cases; the code policies never did. Missing =
normal matched the bounds policy's overall accuracy under noise because
it rarely asks, but when it answered it was right only 89--91\% of the
time, and it under-triaged 8.4--10.2\% of patients in every condition
(the bounds policy: 0.0--0.5\%). With the higher-risk fallback, the
bounds policies over-triaged when information was genuinely unavailable
(9.6--12.9\% under noise), the safe direction. On messy notes, Haiku
extraction lost a little accuracy (98.6\%; bounds + checks recovered
99.1\%), Qwen none (99.5\%), and the agent reading the notes directly
was more accurate than the bounds policy with Haiku (99.8\% vs 98.6\%,
p=0.047).

\needspace{16\baselineskip}

\textbf{Table 1.} Main results with Claude Haiku 4.5 extraction, clean
notes, all 1,200 cases, with an ideal and a noisy clinician. Accuracy
counts abstention as incorrect. Under-triage: a lower-risk category than
the truth, with undetermined cases assigned the highest-risk category
still possible. Results for the oracle and Qwen3.5-9B extractors and for
messy notes are in Supplementary S6.

{\def\LTcaptype{none} 
{\footnotesize\setlength{\tabcolsep}{3pt}
\begin{longtable}[]{@{}
  >{\raggedright\arraybackslash}p{(\linewidth - 12\tabcolsep) * \real{0.1429}}
  >{\raggedright\arraybackslash}p{(\linewidth - 12\tabcolsep) * \real{0.3651}}
  >{\raggedright\arraybackslash}p{(\linewidth - 12\tabcolsep) * \real{0.1111}}
  >{\raggedright\arraybackslash}p{(\linewidth - 12\tabcolsep) * \real{0.0794}}
  >{\raggedright\arraybackslash}p{(\linewidth - 12\tabcolsep) * \real{0.0952}}
  >{\raggedright\arraybackslash}p{(\linewidth - 12\tabcolsep) * \real{0.0952}}
  >{\raggedright\arraybackslash}p{(\linewidth - 12\tabcolsep) * \real{0.1111}}@{}}
\toprule\noalign{}
\begin{minipage}[b]{\linewidth}\raggedright
Clinician
\end{minipage} & \begin{minipage}[b]{\linewidth}\raggedright
Measure
\end{minipage} & \begin{minipage}[b]{\linewidth}\raggedright
Ask-all
\end{minipage} & \begin{minipage}[b]{\linewidth}\raggedright
Agent
\end{minipage} & \begin{minipage}[b]{\linewidth}\raggedright
Bounds
\end{minipage} & \begin{minipage}[b]{\linewidth}\raggedright
Bounds + checks
\end{minipage} & \begin{minipage}[b]{\linewidth}\raggedright
Missing = normal
\end{minipage} \\
\midrule\noalign{}
\endhead
\bottomrule\noalign{}
\endlastfoot
Ideal & Accuracy, \% & 99.4 & 99.6 & 99.4 & 99.7 & 91.2 \\
& Questions per case & 1.78 & 0.99 & 0.92 & 0.90 & 0.21 \\
& Irrelevant questions, \% & 48.1 & 9.5 & 0.0 & 0.0 & 0.0 \\
& Answered too early, \% & 0.0 & 0.2 & 0.0 & 0.0 & 40.0 \\
& Under-triage, \% & 0.1 & 0.1 & 0.1 & 0.1 & 8.5 \\
Noisy & Accuracy, \% & 87.0 & 83.5 & 87.0 & 87.2 & 88.0 \\
& Answered too early, \% & 0.0 & 2.7 & 0.0 & 0.0 & 38.7 \\
& Under-triage, \% & 0.3 & 0.5 & 0.3 & 0.3 & 8.4 \\
\end{longtable}
}
}

\begin{figure}
\centering
\pandocbounded{\includegraphics[keepaspectratio,alt={Decision-category accuracy (left) and questions per case (right) for each policy under the four conditions (Haiku extraction, 559 cases with validated notes in both styles). Each condition has its own lane within a policy's row}]{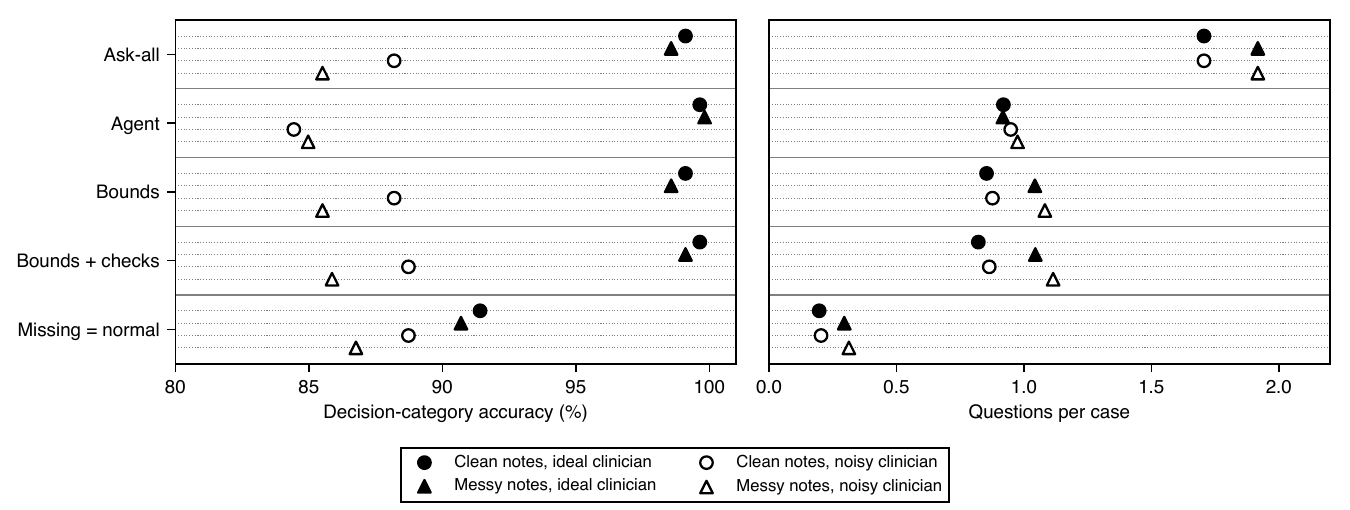}}
\caption{Decision-category accuracy (left) and questions per case
(right) for each policy under the four conditions (Haiku extraction, 559
cases with validated notes in both styles). Each condition has its own
lane within a policy's row}
\end{figure}

\begin{figure}
\centering
\pandocbounded{\includegraphics[keepaspectratio,alt={Under-triage (a lower-risk category than the truth) and over-triage (higher-risk) for each policy under the four conditions, with undetermined cases assigned the highest-risk category still possible (Haiku extraction, 559 paired cases). Symbols and lanes as in Figure 2; horizontal lines are Wilson 95\% confidence intervals}]{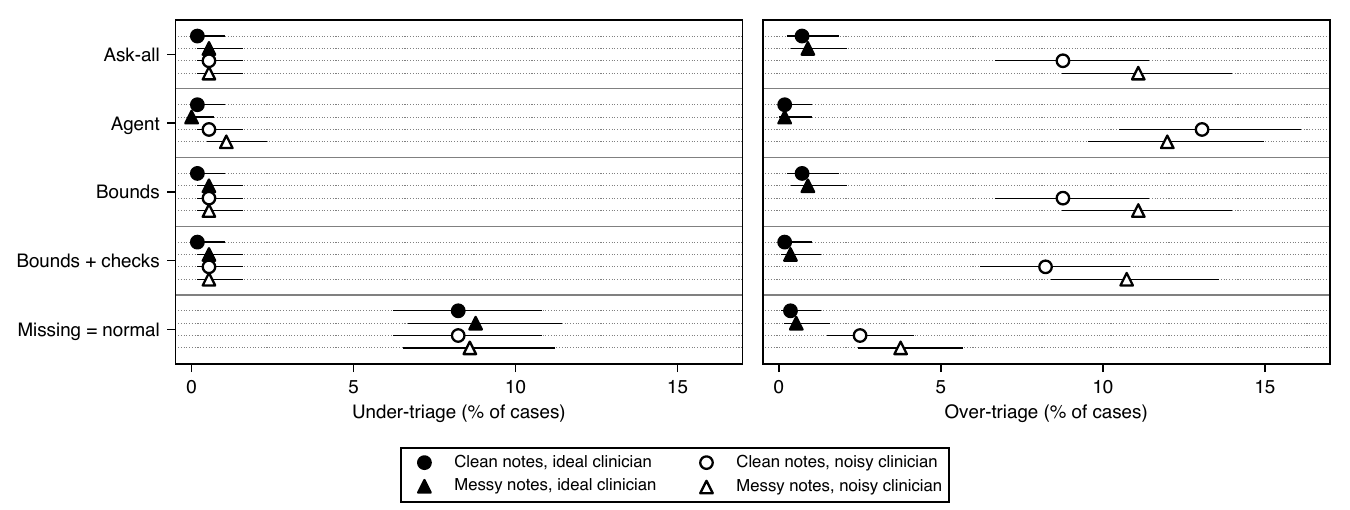}}
\caption{Under-triage (a lower-risk category than the truth) and
over-triage (higher-risk) for each policy under the four conditions,
with undetermined cases assigned the highest-risk category still
possible (Haiku extraction, 559 paired cases). Symbols and lanes as in
Figure 2; horizontal lines are Wilson 95\% confidence intervals}
\end{figure}

\textbf{Real notes (Figure 4).} Our code reproduced MedCalc-Bench's
labels from its annotated inputs for all HEART, CURB-65, PERC and Wells
notes; for Cockcroft-Gault only 51\%, because the benchmark chooses
actual, ideal or adjusted weight by body-mass index. Only 52\% (95\% CI
48--56) of the 584 training case reports had a category determined by
the documented facts (Qwen: 47\%), ranging from 90\% for PERC to 13\%
for HEART. Under the missing-equals-normal convention the category was
correct for 69\% of notes (HEART 22\%). The truth stayed within our
bounds for 93--95\% of notes; the misses were extraction errors, mainly
in HEART inputs. The test split gave similar results (Supplementary S8).

\begin{figure}
\centering
\pandocbounded{\includegraphics[keepaspectratio,alt={Real case reports (MedCalc-Bench Verified, training split; Haiku extraction): share whose category is determined by the note, and share whose category is correct when missing inputs are treated as normal.}]{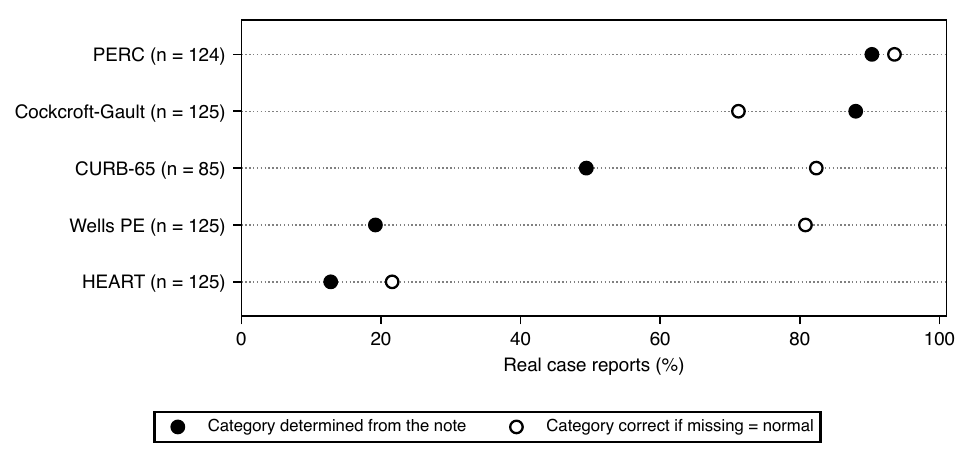}}
\caption{Real case reports (MedCalc-Bench Verified, training split;
Haiku extraction): share whose category is determined by the note, and
share whose category is correct when missing inputs are treated as
normal.}
\end{figure}

\section{4. Discussion}\label{discussion}

The most consistent finding is a safety one: treating undocumented
findings as normal under-triaged about one patient in twelve in every
condition, including clean notes and a perfect clinician, and gave the
right category for only two-thirds of real case reports. Headline
accuracy hides this, because a system that rarely asks rarely meets an
unanswerable question. Letting the model only read, and code decide,
removed premature commitment and irrelevant questions and halved the
questions asked. A frontier agent was a strong baseline, as accurate
under ideal conditions and more accurate than a small extractor on messy
notes, so we do not claim the modular design is more accurate than a
capable agent. Its advantages are that it never answers early, never
asks an irrelevant question, has the lowest under-triage, is
deterministic and auditable, and needs only a model that can read: a 9B
local model matched perfect extraction, whereas the agent needs a
frontier model at every turn.

ClinDet-Bench \citep{watanabe2026clindet} poses the same determinacy
question over score ranges; we add question asking and a comparison with
an agent that can ask. MediQ, AgentClinic and CRAFT-MD evaluate
information seeking in diagnostic dialogue rather than calculators
\citep{li2024mediq, schmidgall2024agentclinic, johri2025craftmd}. Our
real-note analysis suggests that MedCalc-Bench's convention
\citep{khandekar2024medcalc} overstates how often a score can safely be
computed from the note alone. The design relates to active feature
acquisition \citep{saartsechansky2009active}, selective prediction
\citep{geifman2017selective} and partial evaluation
\citep{jones1993partial}, and executable guideline standards such as
Clinical Quality Language \citep{hl7cql} and FHIR Clinical Practice
Guidelines \citep{hl7cpg} are a natural home for the determinacy check.

\textbf{Limitations.} The notes are synthetic, cleaner and more complete
than real ones, and were written by a model from the same family as the
Claude extractor; a small exploratory comparison with Qwen-written notes
suggested a same-family advantage of 3.2 points (95\% CI 0.4--6.4).
Absolute accuracies are therefore optimistic; question counts and safety
differences are the more transferable results. The clinician is
simulated with a simple noise model. Patient characteristics were
sampled independently, which produced 12 implausible HEART cases;
excluding them changed nothing (Supplementary S10). Claude models were
used through headless Claude Code, and their confidence is
self-reported. The real-note analysis could not include question asking,
and its labels have known errors
\citep{ye2025stewardship, krohngrimberghe2026audit}. Calculator readings
and priors were reviewed by one physician. Accuracy is near ceiling
under ideal conditions, so several accuracy comparisons are
underpowered.

\textbf{Implications.} Keep ``unknown'' distinct from ``absent'' end to
end, let code decide whether the documented facts determine the
decision, ask only what can change it, and fall back to the higher-risk
category when information cannot be obtained. Benchmarks that fill
missing inputs with normal values should also report how often the note
actually determines the category.

\section{Declarations}\label{declarations}

\textbf{Data and code.} Code: https://github.com/nicoveraz/calc-bounds
(MIT licence), archived at Zenodo,
https://doi.org/10.5281/zenodo.23004726. The synthetic cohort is
regenerated from the configuration and seed; notes, extractions and the
model-response cache are not released, and re-running regenerates them
(model outputs may differ slightly). MedCalc-Bench Verified is available
from its authors. \textbf{Funding:} none. \textbf{Competing interests:}
none. \textbf{Use of AI:} LLMs generated the synthetic notes and served
as extractors, judge and agent; code and analysis were developed with AI
assistance (Claude Code). \textbf{Ethics:} synthetic data and published
de-identified case reports only; no ethics review was required.

\bibliographystyle{plainnat}
\bibliography{refs}

\clearpage
\appendix

\section{Supplementary material}\label{supplementary-material}

Sections S1 (calculator readings), S6-S7 (full result grids and paired
comparisons) and S9 (exact prompts) are in the full supplementary
material, \texttt{paper/supplement.pdf} in the code repository.

\subsection{S2. Synthetic cohort and population
priors}\label{s2.-synthetic-cohort-and-population-priors}

Cohort (n = 1,200). Undetermined: category not determined by the
documented facts (50\% by design).

{\def\LTcaptype{none} 
{\footnotesize\setlength{\tabcolsep}{3pt}
\begin{longtable}[]{@{}
  >{\raggedright\arraybackslash}p{(\linewidth - 12\tabcolsep) * \real{0.1531}}
  >{\raggedright\arraybackslash}p{(\linewidth - 12\tabcolsep) * \real{0.0510}}
  >{\raggedright\arraybackslash}p{(\linewidth - 12\tabcolsep) * \real{0.1020}}
  >{\raggedright\arraybackslash}p{(\linewidth - 12\tabcolsep) * \real{0.4082}}
  >{\raggedright\arraybackslash}p{(\linewidth - 12\tabcolsep) * \real{0.1224}}
  >{\raggedright\arraybackslash}p{(\linewidth - 12\tabcolsep) * \real{0.1122}}
  >{\raggedright\arraybackslash}p{(\linewidth - 12\tabcolsep) * \real{0.0510}}@{}}
\toprule\noalign{}
\begin{minipage}[b]{\linewidth}\raggedright
Calculator
\end{minipage} & \begin{minipage}[b]{\linewidth}\raggedright
Cases
\end{minipage} & \begin{minipage}[b]{\linewidth}\raggedright
Parameters
\end{minipage} & \begin{minipage}[b]{\linewidth}\raggedright
True categories
\end{minipage} & \begin{minipage}[b]{\linewidth}\raggedright
Undetermined from note
\end{minipage} & \begin{minipage}[b]{\linewidth}\raggedright
Params not documented, \%
\end{minipage} & \begin{minipage}[b]{\linewidth}\raggedright
Cases with traps
\end{minipage} \\
\midrule\noalign{}
\endhead
\bottomrule\noalign{}
\endlastfoot
HEART & 200 & 11 & low 82, moderate 103, high 15 & 100 & 25 & 56 \\
CURB-65 & 200 & 6 & low 95, moderate 68, high 37 & 100 & 24 & 61 \\
qSOFA & 200 & 3 & negative 154, positive 46 & 100 & 33 & 53 \\
PERC & 200 & 8 & negative 86, positive 114 & 100 & 29 & 61 \\
Wells PE & 200 & 7 & pe\_unlikely 162, pe\_likely 38 & 100 & 29 & 56 \\
Cockcroft-Gault & 200 & 4 & crcl\_lt\_30 23, crcl\_30\_to\_lt\_60 91,
crcl\_ge\_60 86 & 100 & 24 & 39 \\
\end{longtable}
}
}

Plausible emergency-department populations for each calculator's
intended use; parameters sampled independently except diastolic
\textless{} systolic blood pressure − 15 mmHg. The PERC cohort
represents low gestalt pre-test probability. Reviewed by the author.

{\def\LTcaptype{none} 
{\footnotesize\setlength{\tabcolsep}{3pt}
\begin{longtable}[]{@{}
  >{\raggedright\arraybackslash}p{(\linewidth - 4\tabcolsep) * \real{0.1875}}
  >{\raggedright\arraybackslash}p{(\linewidth - 4\tabcolsep) * \real{0.3125}}
  >{\raggedright\arraybackslash}p{(\linewidth - 4\tabcolsep) * \real{0.5000}}@{}}
\toprule\noalign{}
\begin{minipage}[b]{\linewidth}\raggedright
Calculator
\end{minipage} & \begin{minipage}[b]{\linewidth}\raggedright
Parameter
\end{minipage} & \begin{minipage}[b]{\linewidth}\raggedright
Prior
\end{minipage} \\
\midrule\noalign{}
\endhead
\bottomrule\noalign{}
\endlastfoot
HEART & heart\_history & slightly\_suspicious 0.45,
moderately\_suspicious 0.35, highly\_suspicious 0.2 \\
HEART & heart\_ecg & normal 0.6, nonspecific\_repolarization 0.3,
significant\_st\_deviation 0.1 \\
HEART & age & Normal(58, 15) truncated to {[}18, 100{]} \\
HEART & hypertension & P(yes) = 0.4 \\
HEART & hypercholesterolemia & P(yes) = 0.3 \\
HEART & diabetes & P(yes) = 0.2 \\
HEART & obesity & P(yes) = 0.25 \\
HEART & smoking & P(yes) = 0.25 \\
HEART & family\_history\_cad & P(yes) = 0.2 \\
HEART & atherosclerotic\_disease & P(yes) = 0.2 \\
HEART & heart\_troponin & le\_normal 0.75, 1\_to\_3x\_normal 0.15,
gt\_3x\_normal 0.1 \\
CURB-65 & confusion & P(yes) = 0.15 \\
CURB-65 & urea & Normal(7, 3.5) truncated to {[}1.5, 40{]} \\
CURB-65 & resp\_rate & Normal(22, 6) truncated to {[}8, 50{]} \\
CURB-65 & sbp & Normal(125, 25) truncated to {[}60, 220{]} \\
CURB-65 & dbp & Normal(72, 14) truncated to {[}30, 130{]} \\
CURB-65 & age & Normal(65, 17) truncated to {[}18, 100{]} \\
qSOFA & resp\_rate & Normal(21, 5) truncated to {[}8, 50{]} \\
qSOFA & altered\_mentation & P(yes) = 0.2 \\
qSOFA & sbp & Normal(118, 24) truncated to {[}60, 220{]} \\
PERC & age & Normal(38, 12) truncated to {[}18, 90{]} \\
PERC & heart\_rate & Normal(88, 15) truncated to {[}45, 160{]} \\
PERC & spo2 & Normal(97, 2) truncated to {[}85, 100{]} \\
PERC & unilateral\_leg\_swelling & P(yes) = 0.05 \\
PERC & hemoptysis & P(yes) = 0.03 \\
PERC & recent\_surgery\_trauma & P(yes) = 0.05 \\
PERC & prior\_vte & P(yes) = 0.05 \\
PERC & hormone\_use & P(yes) = 0.12 \\
Wells PE & dvt\_signs & P(yes) = 0.12 \\
Wells PE & pe\_most\_likely & P(yes) = 0.3 \\
Wells PE & heart\_rate & Normal(95, 18) truncated to {[}45, 170{]} \\
Wells PE & immobilization\_or\_surgery & P(yes) = 0.12 \\
Wells PE & prior\_vte & P(yes) = 0.12 \\
Wells PE & hemoptysis & P(yes) = 0.05 \\
Wells PE & malignancy & P(yes) = 0.1 \\
Cockcroft-Gault & age & Normal(60, 17) truncated to {[}18, 100{]} \\
Cockcroft-Gault & weight & Normal(75, 17) truncated to {[}35, 180{]} \\
Cockcroft-Gault & creatinine & Normal(1.3, 0.8) truncated to {[}0.4,
8{]} \\
Cockcroft-Gault & sex & male 0.5, female 0.5 \\
\end{longtable}
}
}

\subsection{S3. Missingness sensitivity (oracle extraction, natural
share of undetermined
cases)}\label{s3.-missingness-sensitivity-oracle-extraction-natural-share-of-undetermined-cases}

Fresh cohorts (200 cases per calculator) at each missingness level;
stratum results reweighted to the natural share of undetermined cases.

{\def\LTcaptype{none} 
{\footnotesize\setlength{\tabcolsep}{3pt}
\begin{longtable}[]{@{}
  >{\raggedright\arraybackslash}p{(\linewidth - 8\tabcolsep) * \real{0.1803}}
  >{\raggedright\arraybackslash}p{(\linewidth - 8\tabcolsep) * \real{0.2623}}
  >{\raggedright\arraybackslash}p{(\linewidth - 8\tabcolsep) * \real{0.1311}}
  >{\raggedright\arraybackslash}p{(\linewidth - 8\tabcolsep) * \real{0.2295}}
  >{\raggedright\arraybackslash}p{(\linewidth - 8\tabcolsep) * \real{0.1967}}@{}}
\toprule\noalign{}
\begin{minipage}[b]{\linewidth}\raggedright
Missingness
\end{minipage} & \begin{minipage}[b]{\linewidth}\raggedright
System
\end{minipage} & \begin{minipage}[b]{\linewidth}\raggedright
Accuracy (natural share)
\end{minipage} & \begin{minipage}[b]{\linewidth}\raggedright
Questions/case (natural share)
\end{minipage} & \begin{minipage}[b]{\linewidth}\raggedright
Undetermined share
\end{minipage} \\
\midrule\noalign{}
\endhead
\bottomrule\noalign{}
\endlastfoot
0.1 & Ask-all & 0.999 & 0.642 & 0.252 \\
0.1 & Missing = normal & 0.964 & 0.089 & 0.252 \\
0.1 & Bounds & 0.999 & 0.326 & 0.252 \\
0.1 & Bounds + checks & 0.999 & 0.316 & 0.252 \\
0.3 & Ask-all & 0.997 & 1.925 & 0.61 \\
0.3 & Missing = normal & 0.899 & 0.282 & 0.61 \\
0.3 & Bounds & 0.997 & 1.144 & 0.61 \\
0.3 & Bounds + checks & 0.997 & 1.082 & 0.61 \\
0.5 & Ask-all & 0.994 & 3.289 & 0.824 \\
0.5 & Missing = normal & 0.824 & 0.483 & 0.824 \\
0.5 & Bounds & 0.994 & 2.159 & 0.824 \\
0.5 & Bounds + checks & 0.994 & 1.979 & 0.824 \\
\end{longtable}
}
}

\subsection{S4. Confirmation threshold sweep (bounds + checks policy,
clean notes, ideal
clinician)}\label{s4.-confirmation-threshold-sweep-bounds-checks-policy-clean-notes-ideal-clinician}

The bounds + checks policy asks the clinician to confirm a
decision-critical extracted value when its calibrated confidence is
below the threshold.

{\def\LTcaptype{none} 
{\footnotesize\setlength{\tabcolsep}{3pt}
\begin{longtable}[]{@{}
  >{\raggedright\arraybackslash}p{(\linewidth - 10\tabcolsep) * \real{0.1515}}
  >{\raggedright\arraybackslash}p{(\linewidth - 10\tabcolsep) * \real{0.1515}}
  >{\raggedright\arraybackslash}p{(\linewidth - 10\tabcolsep) * \real{0.1212}}
  >{\raggedright\arraybackslash}p{(\linewidth - 10\tabcolsep) * \real{0.2121}}
  >{\raggedright\arraybackslash}p{(\linewidth - 10\tabcolsep) * \real{0.2727}}
  >{\raggedright\arraybackslash}p{(\linewidth - 10\tabcolsep) * \real{0.0909}}@{}}
\toprule\noalign{}
\begin{minipage}[b]{\linewidth}\raggedright
Extractor
\end{minipage} & \begin{minipage}[b]{\linewidth}\raggedright
Confirm below confidence
\end{minipage} & \begin{minipage}[b]{\linewidth}\raggedright
Accuracy
\end{minipage} & \begin{minipage}[b]{\linewidth}\raggedright
Questions/case
\end{minipage} & \begin{minipage}[b]{\linewidth}\raggedright
Confirmations/case
\end{minipage} & \begin{minipage}[b]{\linewidth}\raggedright
Errors caught
\end{minipage} \\
\midrule\noalign{}
\endhead
\bottomrule\noalign{}
\endlastfoot
Haiku 4.5 & 0.5 & 0.995 & 0.8833 & 0.005 & 5 \\
Haiku 4.5 & 0.8 & 0.9967 & 0.895 & 0.015 & 9 \\
Haiku 4.5 & 0.9 & 0.9967 & 0.895 & 0.015 & 9 \\
Haiku 4.5 & 0.95 & 0.9967 & 0.895 & 0.015 & 9 \\
Haiku 4.5 & 0.99 & 0.9983 & 1.705 & 0.8242 & 19 \\
Haiku 4.5 & 0.999 & 0.9983 & 1.7075 & 0.8267 & 19 \\
Qwen3.5-9B & 0.5 & 0.9983 & 1.1783 & 0.0 & 0 \\
Qwen3.5-9B & 0.8 & 0.9983 & 1.1783 & 0.0 & 0 \\
Qwen3.5-9B & 0.9 & 0.9983 & 1.1983 & 0.02 & 2 \\
Qwen3.5-9B & 0.95 & 0.9983 & 1.2967 & 0.1183 & 5 \\
Qwen3.5-9B & 0.99 & 0.9983 & 1.8142 & 0.6358 & 6 \\
Qwen3.5-9B & 0.999 & 0.9983 & 1.8142 & 0.6358 & 6 \\
\end{longtable}
}
}

\subsection{S5. Calibration of extraction
confidence}\label{s5.-calibration-of-extraction-confidence}

Present/absent claims; isotonic and temperature calibration fitted on a
seeded 30\% development split, evaluated on the rest. Haiku 4.5
(self-reported): ECE 0.014 raw, 0.003 temperature, 0.002 isotonic.
Qwen3.5-9B (token log-probabilities): ECE 0.048 raw, 0.033 temperature,
0.002 isotonic.

\begin{figure}
\centering
\pandocbounded{\includegraphics[keepaspectratio,alt={Reliability of raw extraction confidence (test split).}]{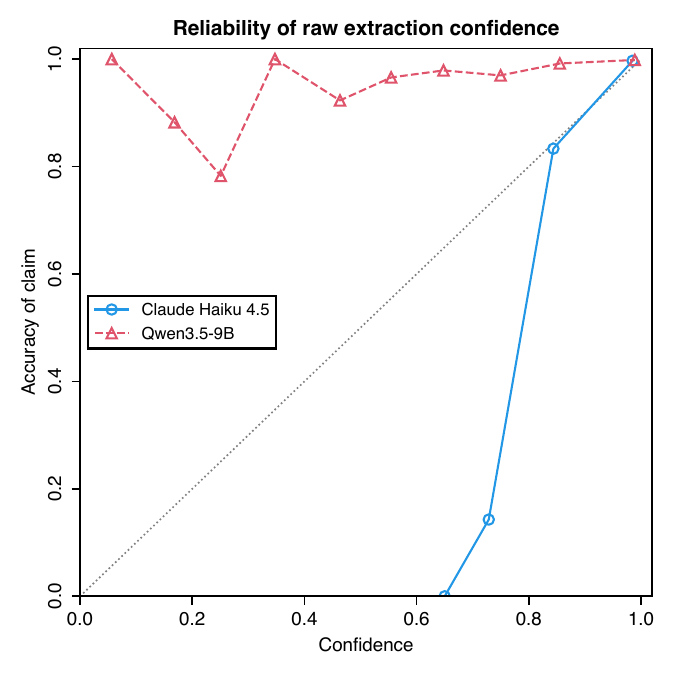}}
\caption{Reliability of raw extraction confidence (test split).}
\end{figure}

\subsection{S8. Real case reports, all splits and
extractors}\label{s8.-real-case-reports-all-splits-and-extractors}

{\def\LTcaptype{none} 
{\footnotesize\setlength{\tabcolsep}{3pt}
\begin{longtable}[]{@{}
  >{\raggedright\arraybackslash}p{(\linewidth - 16\tabcolsep) * \real{0.0532}}
  >{\raggedright\arraybackslash}p{(\linewidth - 16\tabcolsep) * \real{0.1064}}
  >{\raggedright\arraybackslash}p{(\linewidth - 16\tabcolsep) * \real{0.1596}}
  >{\raggedright\arraybackslash}p{(\linewidth - 16\tabcolsep) * \real{0.0426}}
  >{\raggedright\arraybackslash}p{(\linewidth - 16\tabcolsep) * \real{0.1064}}
  >{\raggedright\arraybackslash}p{(\linewidth - 16\tabcolsep) * \real{0.1702}}
  >{\raggedright\arraybackslash}p{(\linewidth - 16\tabcolsep) * \real{0.0957}}
  >{\raggedright\arraybackslash}p{(\linewidth - 16\tabcolsep) * \real{0.1596}}
  >{\raggedright\arraybackslash}p{(\linewidth - 16\tabcolsep) * \real{0.1064}}@{}}
\toprule\noalign{}
\begin{minipage}[b]{\linewidth}\raggedright
Split
\end{minipage} & \begin{minipage}[b]{\linewidth}\raggedright
Extractor
\end{minipage} & \begin{minipage}[b]{\linewidth}\raggedright
Calculator
\end{minipage} & \begin{minipage}[b]{\linewidth}\raggedright
n
\end{minipage} & \begin{minipage}[b]{\linewidth}\raggedright
Code reproduces label, \%
\end{minipage} & \begin{minipage}[b]{\linewidth}\raggedright
Determined from note, \% (95\% CI)
\end{minipage} & \begin{minipage}[b]{\linewidth}\raggedright
Truth still possible, \%
\end{minipage} & \begin{minipage}[b]{\linewidth}\raggedright
Category correct, missing=normal, \%
\end{minipage} & \begin{minipage}[b]{\linewidth}\raggedright
Entity agreement, \%
\end{minipage} \\
\midrule\noalign{}
\endhead
\bottomrule\noalign{}
\endlastfoot
train & Haiku 4.5 & Cockcroft-Gault & 125 & 51 & 88.0 (81.1--92.6) & 85
& 71 & 90 \\
train & Haiku 4.5 & CURB-65 & 85 & 100 & 49.4 (39.0--59.8) & 93 & 82 &
89 \\
train & Haiku 4.5 & HEART & 125 & 100 & 12.8 (8.0--19.8) & 90 & 22 &
73 \\
train & Haiku 4.5 & PERC & 124 & 100 & 90.3 (83.8--94.4) & 100 & 94 &
90 \\
train & Haiku 4.5 & Wells PE & 125 & 100 & 19.2 (13.3--27.0) & 98 & 81 &
87 \\
train & Haiku 4.5 & All & 584 & 90 & 52.1 (48.0--56.1) & 93 & 69 & 86 \\
train & Qwen3.5-9B & Cockcroft-Gault & 125 & 51 & 84.8 (77.5--90.0) & 90
& 72 & 92 \\
train & Qwen3.5-9B & CURB-65 & 85 & 100 & 41.2 (31.3--51.8) & 95 & 81 &
92 \\
train & Qwen3.5-9B & HEART & 125 & 100 & 12.0 (7.4--18.9) & 93 & 9 &
74 \\
train & Qwen3.5-9B & PERC & 124 & 100 & 87.9 (81.0--92.5) & 100 & 91 &
89 \\
train & Qwen3.5-9B & Wells PE & 125 & 100 & 9.6 (5.6--16.0) & 99 & 78 &
88 \\
train & Qwen3.5-9B & All & 584 & 90 & 47.4 (43.4--51.5) & 95 & 65 &
87 \\
test & Haiku 4.5 & Cockcroft-Gault & 20 & 40 & 95.0 (76.4--99.1) & 100 &
95 & 96 \\
test & Haiku 4.5 & CURB-65 & 20 & 100 & 55.0 (34.2--74.2) & 100 & 100 &
97 \\
test & Haiku 4.5 & HEART & 20 & 95 & 60.0 (38.7--78.1) & 85 & 45 & 87 \\
test & Haiku 4.5 & PERC & 20 & 100 & 75.0 (53.1--88.8) & 100 & 95 &
99 \\
test & Haiku 4.5 & Wells PE & 20 & 100 & 10.0 (2.8--30.1) & 100 & 85 &
89 \\
test & Haiku 4.5 & All & 100 & 87 & 59.0 (49.2--68.1) & 97 & 84 & 94 \\
test & Qwen3.5-9B & Cockcroft-Gault & 20 & 40 & 90.0 (69.9--97.2) & 100
& 90 & 96 \\
test & Qwen3.5-9B & CURB-65 & 20 & 100 & 50.0 (29.9--70.1) & 95 & 90 &
98 \\
test & Qwen3.5-9B & HEART & 20 & 95 & 30.0 (14.5--51.9) & 95 & 35 &
77 \\
test & Qwen3.5-9B & PERC & 20 & 100 & 75.0 (53.1--88.8) & 100 & 95 &
96 \\
test & Qwen3.5-9B & Wells PE & 20 & 100 & 10.0 (2.8--30.1) & 100 & 85 &
89 \\
test & Qwen3.5-9B & All & 100 & 87 & 51.0 (41.3--60.6) & 98 & 79 & 91 \\
\end{longtable}
}
}

\subsection{S10. Sensitivity to implausible synthetic
cases}\label{s10.-sensitivity-to-implausible-synthetic-cases}

Characteristics were sampled independently, so some cases are clinically
implausible. We flagged established atherosclerotic disease with no risk
factors, and age under 40 with three or more risk factors or
atherosclerotic disease. We then recomputed the main results without
those cases (clean notes, ideal clinician).

12 of 1200 cases flagged: heart-0017 (age \textless{} 40 with
\textgreater= 3 risk factors or atherosclerotic disease); heart-0023
(age \textless{} 40 with \textgreater= 3 risk factors or atherosclerotic
disease); heart-0036 (age \textless{} 40 with \textgreater= 3 risk
factors or atherosclerotic disease); heart-0069 (age \textless{} 40 with
\textgreater= 3 risk factors or atherosclerotic disease); heart-0087
(established atherosclerotic disease with no risk factors); heart-0104
(established atherosclerotic disease with no risk factors); heart-0128
(established atherosclerotic disease with no risk factors); heart-0132
(established atherosclerotic disease with no risk factors); heart-0135
(age \textless{} 40 with \textgreater= 3 risk factors or atherosclerotic
disease); heart-0137 (established atherosclerotic disease with no risk
factors); heart-0172 (age \textless{} 40 with \textgreater= 3 risk
factors or atherosclerotic disease); heart-0195 (established
atherosclerotic disease with no risk factors)

{\def\LTcaptype{none} 
{\footnotesize\setlength{\tabcolsep}{3pt}
\begin{longtable}[]{@{}
  >{\raggedright\arraybackslash}p{(\linewidth - 12\tabcolsep) * \real{0.1205}}
  >{\raggedright\arraybackslash}p{(\linewidth - 12\tabcolsep) * \real{0.2048}}
  >{\raggedright\arraybackslash}p{(\linewidth - 12\tabcolsep) * \real{0.1928}}
  >{\raggedright\arraybackslash}p{(\linewidth - 12\tabcolsep) * \real{0.0482}}
  >{\raggedright\arraybackslash}p{(\linewidth - 12\tabcolsep) * \real{0.1084}}
  >{\raggedright\arraybackslash}p{(\linewidth - 12\tabcolsep) * \real{0.1566}}
  >{\raggedright\arraybackslash}p{(\linewidth - 12\tabcolsep) * \real{0.1687}}@{}}
\toprule\noalign{}
\begin{minipage}[b]{\linewidth}\raggedright
Extraction
\end{minipage} & \begin{minipage}[b]{\linewidth}\raggedright
Cases
\end{minipage} & \begin{minipage}[b]{\linewidth}\raggedright
System
\end{minipage} & \begin{minipage}[b]{\linewidth}\raggedright
n
\end{minipage} & \begin{minipage}[b]{\linewidth}\raggedright
Accuracy, \%
\end{minipage} & \begin{minipage}[b]{\linewidth}\raggedright
Under-triage, \%
\end{minipage} & \begin{minipage}[b]{\linewidth}\raggedright
Questions/case
\end{minipage} \\
\midrule\noalign{}
\endhead
\bottomrule\noalign{}
\endlastfoot
Oracle & all & Ask-all & 1200 & 99.8 & 0.0 & 1.74 \\
Oracle & all & Agent & 1200 & 99.6 & 0.1 & 0.99 \\
Oracle & all & Bounds & 1200 & 99.8 & 0.0 & 0.92 \\
Oracle & all & Bounds + checks & 1200 & 99.8 & 0.0 & 0.87 \\
Oracle & all & Missing = normal & 1200 & 91.8 & 8.2 & 0.21 \\
Oracle & excluding flagged & Ask-all & 1188 & 99.8 & 0.0 & 1.73 \\
Oracle & excluding flagged & Agent & 1188 & 99.6 & 0.1 & 0.99 \\
Oracle & excluding flagged & Bounds & 1188 & 99.8 & 0.0 & 0.92 \\
Oracle & excluding flagged & Bounds + checks & 1188 & 99.8 & 0.0 &
0.87 \\
Oracle & excluding flagged & Missing = normal & 1188 & 91.9 & 8.1 &
0.21 \\
Haiku 4.5 & all & Ask-all & 1200 & 99.4 & 0.1 & 1.78 \\
Haiku 4.5 & all & Agent & 1200 & 99.6 & 0.1 & 0.99 \\
Haiku 4.5 & all & Bounds & 1200 & 99.4 & 0.1 & 0.92 \\
Haiku 4.5 & all & Bounds + checks & 1200 & 99.7 & 0.1 & 0.90 \\
Haiku 4.5 & all & Missing = normal & 1200 & 91.2 & 8.5 & 0.21 \\
Haiku 4.5 & excluding flagged & Ask-all & 1188 & 99.4 & 0.1 & 1.77 \\
Haiku 4.5 & excluding flagged & Agent & 1188 & 99.6 & 0.1 & 0.99 \\
Haiku 4.5 & excluding flagged & Bounds & 1188 & 99.4 & 0.1 & 0.93 \\
Haiku 4.5 & excluding flagged & Bounds + checks & 1188 & 99.7 & 0.1 &
0.90 \\
Haiku 4.5 & excluding flagged & Missing = normal & 1188 & 91.3 & 8.3 &
0.21 \\
Qwen3.5-9B & all & Ask-all & 1200 & 99.8 & 0.0 & 2.21 \\
Qwen3.5-9B & all & Agent & 1200 & 99.6 & 0.1 & 0.99 \\
Qwen3.5-9B & all & Bounds & 1200 & 99.8 & 0.0 & 1.23 \\
Qwen3.5-9B & all & Bounds + checks & 1200 & 99.8 & 0.0 & 1.20 \\
Qwen3.5-9B & all & Missing = normal & 1200 & 91.2 & 8.8 & 0.26 \\
Qwen3.5-9B & excluding flagged & Ask-all & 1188 & 99.8 & 0.0 & 2.20 \\
Qwen3.5-9B & excluding flagged & Agent & 1188 & 99.6 & 0.1 & 0.99 \\
Qwen3.5-9B & excluding flagged & Bounds & 1188 & 99.8 & 0.0 & 1.23 \\
Qwen3.5-9B & excluding flagged & Bounds + checks & 1188 & 99.8 & 0.0 &
1.20 \\
Qwen3.5-9B & excluding flagged & Missing = normal & 1188 & 91.4 & 8.6 &
0.26 \\
\end{longtable}
}
}

\end{document}